\documentclass{article}

\usepackage[english]{babel}

\usepackage[letterpaper,top=2cm,bottom=2cm,left=3cm,right=3cm,marginparwidth=1.75cm]{geometry}

\usepackage{amsmath}
\usepackage{csquotes}
\usepackage{graphicx}
\usepackage[colorlinks=true, allcolors=blue]{hyperref}

\title{Witgenstein's influence on artificial intelligence}
\author{Piero Molino and Jacopo Tagliabue\footnote{This is an English pre-print of the chapter that appeared in Spanish in \textit{CENTENARIO DEL SILENCIO}. At the time of submitting our work (2021), Piero Molino was at Stanford University and Jacopo Tagliabue at Coveo Labs: the current text has been only slightly edited for content and citations.}}

\begin{document}
\maketitle

\begin{abstract}

We examine how much of the contemporary progress in artificial intelligence (and, specifically, in natural language processing), can be, more or less directly, traced back to the seminal work and ideas of the Austrian-British philosopher Ludwig Wittgenstein, with particular focus on his late views. Discussing Wittgenstein's original theses will give us the chance to survey the state of artificial intelligence, and comment on both its strengths and weaknesses. A similar text appeared first in Spanish as a chapter of \textit{CENTENARIO DEL SILENCIO} (2021), a book celebrating 100 years since the publication of the \textit{Tractatus}. 

\end{abstract}

The thesis we are proposing here is one that is not actively debated in artificial intelligence research circles to the best of our knowledge. However, we believe it could help interpret recent lines of research that have proved themselves effective. The thesis is that much of the progress in artificial intelligence, and specifically in natural language processing and machine learning in recent years can be, albeit indirectly, traced back to Witgenstein, specifically to the ``Philosophical Investigations'' (henceforth, \textit{PI}). In some instances the connection is easier to trace, in some others there is a leap to take in order to accept the thesis, but despite that, we believe exploring it could be beneficial, even if only for the inspirations it could trigger by analogy.

We suggest following the thread of research on natural language processing that more directly follows from Witgenstein through the work of Margaret Masterman as it will lead to an intersection point with machine learning later on. Margaret Masterman was a student in Wittgenstein's 1933-1934 course, the lecture notes of which were collected in the ``Blue Book''. The ideas in that course were the foundational ones later contained in the ``Philosophical Investigations'': the specific insight that the meaning of words derives from their use in language constituted the basis of Masterman's computational approach to linguistics - focused on meaning and thesauri rather than syntax -, which she applied in particular to machine translation. She founded the Cambridge Language Research Unit in 1955, and work coming from there influenced researchers all around the world for the following decades, in a time when Chomskian ideas \cite{reason:Chomsky57a} were predominant (Sixties) and Montague proposed his logic-based theories \cite{montague1974d} (early Seventies).

Around the same time as Masterman, Zellig Harris was publishing his most influential structuralist works \cite{Harris1951,harris54}. They followed in the American linguistic tradition of Leonard Bloomfield, but in fact echoed the structuralist theories coming from Cambridge. With ``Papers in Linguistics'' \cite{Firth_Papers57} and ``Mathematical Structure of Language'' \cite{harris1968} we have some of the first appearances of the distributional hypothesis, which states that the degree of semantic similarity between two linguistic expressions is a function of the similarity of the linguistic contexts in which they can appear.

This hypothesis has been the foundation of countless empirical and computational works in the sub-field of natural language processing called distributional semantics (for a detailed historical account see \cite{doi:10.1146/annurev-linguistics-030514-125254}). In short, many different algorithms have been proposed to represent words as vectors in a multi-dimensional space, usually by statistical methods observing the co-occurrence of words with other words or other representations of contexts within a specific corpus. The vectors obtained this way have interesting properties that follow the distributional hypothesis and have been fruitfully applied in many applications, particularly in the context of information retrieval and search engines starting in the 70s with the vector space model up to the current day.

A very influential work in 2013 \cite{NIPS2013_9aa42b31} proposed \textit{word2vec} \footnote{Technically \textit{word2vec} is a tool containing two algorithmic approaches: \textit{continuous bag of words} and \textit{skip gram}.}, a novel algorithm to obtain vectors representing words. The algorithms consist of a function that predicts a word given its context (continuous bag of words) or the context given a word (skip gram), where words and contexts are n-grams obtained from a large corpus. For instance, first a text like ``The quick brown fox jumps over the lazy dog'' is transformed in n-grams of consecutive words like ``The quick brown fox jumps'', ``quick brown fox jumps over'', ``brown fox jumps over the'' and so on, then a function that computes $P(word|context)$ is trained to maximize the probability of ``brown'' given ``The quick fox jumps'', and the process is repeated for all the other n-grams in the corpus. The function is parametrized by a shallow neural network made of only two linear layers and its parameters are learned through optimization (stochastic gradient descent). Importantly, the learned parameters of the first layer are then used as vector representations of the words in the corpus.  One of the interesting properties of the vector spaces obtained with this method is that some relationships between words are captured as linear directions in the space, making it possible to use linear algebra for simple reasoning, like the famous example ``king is to man what queen is to ...'' that can be solved by calculating the closest vector to ``vec(king) - vec(man) + vec(queen)''.

Despite the simplicity and the fact that fundamentally its workings are no more than a direct implementation of structuralist principles, this algorithm changed the perception of the entire field, as it brought machine learning and distributional semantics communities much closer.
Its success is also due to the timeliness of proposing a method to obtain distributed representations of words well suited for use with neural networks, making it possible to adapt deep learning approaches to textual data, in a moment when deep learning was successfully applied mostly to other types of perceptual data. \textit{Word2vec} was not the first neural network approach to obtain vector representations though, but it was the first one to popularize the idea, because it was practical, fast and easy to use. \cite{10.5555/944919.944966} proposed a neural network algorithm that considered only the words preceding a certain word in order to predict the following one, what is commonly referred to as language modeling, but they did not use dense word representations, which made the approach impractical. \cite{Miikkulainen1988FormingGR} very pionieristically proposed the FGREP algorithm, a kind of auto-encoder method that obtained representations by means of compressing and predicting back sentences and semantic frames, clearly ahead of its time.

After studies on strengths and limitations of learning-based algorithms as opposed to classical distributional semantics ones \cite{baroni-etal-2014-dont,levy-etal-2015-improving,sahlgren-lenci-2016-effects}, the natural language processing community accepted machine learning as the \textit{de facto} standard way to obtain word representations. Since then many other neural network based approaches have been proposed as alternatives, with one of the most widely applied today being GloVe \cite{pennington-etal-2014-glove}, and those algorithms have been applied for all sorts of natural language processing tasks, and have influenced new developments in recommender systems, graph algorithms, speech recognition and many more.

The reason they work is the basic structuralist assumption that meaning emerges by use and is determined by context. The reason why structuralist assumptions are so important is because of the cost of acquiring data.
The most widely adopted paradigm in machine learning is supervised learning, which consists in having humans tag or categorize objects and then iteratively improving an algorithm by means of how well its predictions match the ones provided by humans. Having humans annotate all the objects of interest is very expensive and error prone, and the most accurate supervised algorithms are data hungry. The famous ImageNet competition \cite{5206848} collected a million images and that was a sufficiently large amount of data for the most effective models to learn to recognize objects in images with high accuracy, but was also really expensive to collect. The structuralist principle instead suggests that meaning depends on use, and as such, you just need to observe use to obtain representations. \textit{Word2vec} and similar algorithms use huge amounts of textual data, but no human labels are provided, the words to predict are simply removed from the ngrams collected, and unlabeled data is obviously extremely cheaper to collect, and it works well because applied well known and very effective supervised learning techniques to huge amounts of inexpensive unsupervised data. In recent years this idea of removing something from the context it appears in and predicting it back has been named self-supervision and has been one of the main focus of research and source of advancement in both natural language processing and machine learning.

Specifically in the natural language processing space, algorithms that learn to represent entire sentences by training on unsupervised corpora have been shown to be very effective when then applied and adapted to supervised tasks. Algorithms like \textit{ELMo} \cite{peters-etal-2018-deep}, \textit{BERT} \cite{devlin-etal-2019-bert} and their many derivatives work exactly in this way. \textit{BERT} for instance is trained with the so-called masked language model objective, which consists in obtaining a sentence from a corpus, masking a few random words and then asking the neural network to predict the masked words from the unmasked ones. This allows training on huge amounts of text crawled from the web and is very effective. At the same time, generative language models, like the \textit{GPT} \cite{Radford2018ImprovingLU} family of algorithms, in which every word in a document is predicted from the words that preceded it, have been shown to be really effective at learning how to generate text that is fluent, usually grammatical, and usually thematically coherent with what came before.

In the broader machine learning field, self-supervision is attracting more and more the interest of the research community. In computer vision, autoencoding and in-filling tasks (in which an area of an image is masked and the algorithm is asked to predict it using the unmasked portion as input) are used and self-supervision is quickly becoming the most effective way to obtain pre-trained models \cite{Pathak2016ContextEF}.
In video analysis, prediction of the next frame is a widely used self-supervision task \cite{Wang2015UnsupervisedLO}, and in graph learning, representations of nodes in graphs are obtained by removing a node from its neighborhood and predicting it back \cite{Grover2016node2vecSF}. Finally, in reinforcement learning, researchers are learning models of the dynamics of the environment in which agents interact by means of observing sequences of actions and states, removing the last one and predicting it given the previous ones, in a similar way to language models \cite{Ha2018WorldM}.

Notwithstanding the influence on contemporary distributional semantics that we just outlined, we do believe that the ``meaning is use''-thesis at the heart of the “Philosophical Investigations” still has insights waiting to be fully appreciated by mainstream Natural Language Processing.  As parting thoughts, we highlight here two broad areas of inquiry for which Wittgenstein's philosophical work may provide interesting intuitions.

The first topic is the relation between words and objects. While it is true that \textit{word2vec} models meaning as a function of linguistic context, it is important to remember that Wittgenstein’s main foe in the PI was his younger self, as well as the entire tradition of scholars analyzing human languages with formal tools (i.e. Tarski-like semantics \cite{Tarski1936-TARTCO,Davidson1967-DAVTAM-3}). When Wittgenstein stresses the importance of context is mostly to reject the idea of a one-to-one correspondence between words and objects, but he certainly does not imply that language is entirely self-referential and can be analyzed independently from reality, like \textit{word2vec} and \textit{BERT} do. To the contrary, even the simpler example of ``language game'' involves an inextricable mixture of linguistic signs and extra-linguistic activity:

\begin{displayquote}
The language is meant to serve for communication between a builder A and an assistant B. A is building with building-stones: there are blocks, pillars, slabs and beams. B has to pass the stones, in the order in which A needs them. For this purpose they use a language consisting of the words "block", "pillar" "slab", "beam". A calls them out; — B brings the stone which he has learnt to bring at such-and-such a call. Conceive this as a complete primitive language. (PI 2.)
\end{displayquote}

While most of the practical success of recent NLP is due to intra-language models, the community is starting to recognize the importance of grounding language in an extra-linguistic reality \cite{bisk-etal-2020-experience,bianchi-etal-2021-language}. Applying the PI lesson to this growing field means not only recognizing the importance of the outside world for the concept of meaning, but also trying to understand communication in the context of human activities: formal studies tend to focus on descriptive language (e.g. Wikipedia pages), but PI made clear that possibly the majority of language use is non-descriptive \cite{austin1962how}: promises, greetings, jokes, etc.. How to account for the variety of intents of human language is an open research question that likely involves modelling social and psychological traits. 

The value of practical communication above descriptive sentences and the importance of social interactions for meaning bring us to the second topic. While Wittgenstein was not particularly fond of mental states, it is true that the PI marked a discontinuity in philosophy of language: if formal accounts mostly assumed that meaning was by and large explicitly coded in linguistic expressions, ``meaning is use'' highlighted that in many cases what is said is significantly less/different than what is meant - semantics is important, but so is pragmatics. Some twenty years after PI, Paul Grice \cite{grice1975logic} put forward a theory of meaning based on “presuppositions”, that is, a framework in which meaning is heavily dependent on context via the expectations of the participants in a conversation. If Richard Montague is mostly worried about logical implication \cite{Montague:1973}, Grice studies \textit{implicatures}: if Jack writes a supporting letter to a Ph.D. candidate, Bob, saying he is a very pleasant person, the committee will most likely think that Jack does \textit{not} believe Bob is very talented, or otherwise he would have said that first. The extent to which Grice's rules are necessary to account for empirical phenomena is subject to debate, but there is general consensus that successful communication involves at least partially the theory of mind of the participants. To the best of our knowledge, no practical NLP application has been so far grounded in formal pragmatics, but in recent years the Rational Speech Act \cite{Goodman2016PragmaticLI} theory has made some progress (mostly in toy examples) in translating Grice's insights into a unified Bayesian framework for linguistic inference. Nested Bayesian inference is not without practical challenges and it is too early to judge how influential this thread of research will be on mainstream NLP; however, RSA is a good example of how formal tooling from statistics and computer science can be used to make philosophical intuitions precise, and, in turn, raise new philosophical questions.

This historical \textit{excursus} followed by a brief description of the current state of artificial intelligence research in many of its sub-fields was aimed at trying to draw a \textit{fil rouge} that can trace back the underlying (and in most cases unconscious and un-acknowledged) assumptions of current work in self-supervision all the way back to structuralism and the theories of Witgenstein and this hopefully also makes the case for researchers in artificial intelligence to be curious about philosophy and its history and to be thought provoking for philosophers with regards to the practical computational implications of their theories. 

It is unlikely that a single discipline, in isolation, will solve the mystery of language; multi-disciplinary, bold thoughts are needed to make significant progress. To quote Ludwig one last time: 

\begin{displayquote}
You could attach prices to thoughts. Some cost a lot, some a little. And how does one pay for thoughts? The answer, I think, is: with courage.
\end{displayquote}

\bibliographystyle{alpha}
\bibliography{sample}

\end{document}